%% file: main.tex
\documentclass[lettersize,journal]{IEEEtran}
\usepackage{amsmath,amsfonts}
\usepackage{algorithmic}
\usepackage{array}
\usepackage[caption=false,font=normalsize,labelfont=sf,textfont=sf]{subfig}
\usepackage{textcomp}
\usepackage{stfloats}
\usepackage{url}
\usepackage{verbatim}
\usepackage{graphicx}
\usepackage{subfig}
\def\BibTeX{{\rm B\kern-.05em{\sc i\kern-.025em b}\kern-.08em
    T\kern-.1667em\lower.7ex\hbox{E}\kern-.125emX}}
\usepackage{balance}
\usepackage[nolist]{acronym}
\usepackage{enumitem}
\usepackage{booktabs}
\usepackage{multirow}
\usepackage{cite}

\newcommand{\eg}{e.g., }

\newcommand{\ie}{i.e., }

\newcommand{\fig}[1]{Fig.~}
\newcommand{\tab}[1]{Tab.~}
\newcommand{\Sec}[1]{Sec.~}
\newcommand{\eq}[1]{Eq.~}

\begin{document}
\include{acro}
\title{SPECI: Skill Prompts based Hierarchical Continual Imitation Learning for Robot Manipulation}
\author{Jingkai Xu, Xiangli Nie \thanks{This work was partly supported by the National Key Research and Development Program of China under Grant 2024YFB4709100, and partly sponsored by Beijing Nova Program under Grants 20220484070 and 20240484498. (Corresponding author: Xiangli Nie).}
\thanks{J. Xu and X. Nie are with the State Key Laboratory of Multimodal Artificial Intelligence Systems, Institute of Automation, Chinese Academy of Sciences, Beijing 100190, China, and also with the School of Artificial Intelligence, University of Chinese Academy of Sciences, Beijing 100049, China. (email: xujingkai2024@ia.ac.cn; xiangli.nie@ia.ac.cn)}
}

\maketitle

\begin{abstract} 
Real-world robot manipulation in dynamic unstructured environments requires lifelong adaptability to evolving objects, scenes and tasks. Traditional imitation learning relies on static training paradigms, which are ill-suited for lifelong adaptation. Although \ac{CIL} enables incremental task adaptation while preserving learned knowledge, current \ac{CIL} methods primarily overlook the intrinsic skill characteristics of robot manipulation or depend on manually defined and rigid skills, leading to suboptimal cross-task knowledge transfer. 
To address these issues, we propose Skill Prompts-based HiErarchical Continual Imitation Learning (SPECI), a novel end-to-end hierarchical \ac{CIL} policy architecture for robot manipulation. The SPECI framework consists of a multimodal perception and fusion module for heterogeneous sensory information encoding, a high-level skill inference module for dynamic skill extraction and selection, and a low-level action execution module for precise action generation. To enable efficient knowledge transfer on both skill and task levels, SPECI performs continual implicit skill acquisition and reuse via an expandable skill codebook and an attention-driven skill selection mechanism. 
Furthermore, we introduce mode approximation to augment the last two modules with task-specific and task-sharing parameters, thereby enhancing task-level knowledge transfer. 
Extensive experiments on diverse manipulation task suites demonstrate that SPECI consistently outperforms state-of-the-art \ac{CIL} methods across all evaluated metrics, revealing exceptional  bidirectional knowledge transfer and superior overall performance.
\end{abstract}

\begin{IEEEkeywords}
Robot manipulation, continual imitation learning, skill acquisition, hierarchical policy, knowledge transfer.
\end{IEEEkeywords}

\section{Introduction}
\IEEEPARstart{D}{eveloping} a general robotic agent capable of performing a wide range of manipulation tasks in unstructured and dynamic environments is the ultimate goal of robot learning. To achieve this ambitious objective, the robotic agent must demonstrate hierarchical reasoning capabilities: environmental understanding at the task level, motion planning at the skill level, and precise execution at the action level.
\ac{IL} and \ac{RL} are two dominant paradigms in robot learning for manipulation tasks. Mainstream RL-based policies require a suitable reward function for every task, which is often hard to design.  Furthermore, RL agents may inadvertently produce unintended and potentially dangerous actions \cite{turner2020avoiding}. In contrast, \ac{IL} is a supervised learning method that extracts a policy by mimicking the actions of a human expert through demonstrations \cite{osa2018algorithmic}. This approach allows researchers to train robotic agents for various manipulation tasks without explicit programming \cite{billard2016learning}.  
Consequently, IL has gained increasing attention in recent research \cite{hao2021sozil, kulak2021combining, brohan2023rt}. However, these methods remain susceptible to failure in complex long-horizon tasks due to accumulated errors arising from naive end-to-end learning \cite{ross2011reduction}.  

To enable \ac{IL} to handle complex manipulation tasks, Zhang et al. \cite{zhang2021explainable} introduced a task planner for high-level decomposition and an \ac{IL} policy for plan-conditioned action execution.  Although this approach demonstrates improved task completion rates, its sequential sub-task paradigm is inherently susceptible to transition errors \cite{huang2023skill}, wherein state misalignment between successive sub-tasks frequently causes execution failures. 
Subsequent research has shifted focus toward skill primitive selection and execution frameworks, which facilitate more granular transitions and enable skill reuse. Nevertheless, these methodologies exhibit two fundamental limitations: either they rely on inflexible, non-adaptive primitive libraries \cite{luo2024multi}, \cite{huang2023skill}, \cite{du2022learning}, or they struggle to determine appropriate levels of skill abstraction through manual specification \cite{wang2023task}, \cite{belkhale2024rt}. Recognizing the inherent complexity of state-action spaces and the challenges associated with explicit behavior quantification, recent works \cite{wang2023mimicplay, xu2023xskill, wang2024novel, mete2024quest, zheng2024prise} have adopted data-driven skill discovery to address these limitations. However, these approaches remain constrained by their dependence on fixed task sets, while real-world robotic agents demand continual adaptation to novel tasks that cannot be fully anticipated during initial deployment \cite{tavassoli2023learning}. Although online fine-tuning with newly collected data offers a potential solution, this strategy is fundamentally limited by the critical challenge of catastrophic forgetting in deep neural networks \cite{mccloskey1989catastrophic}.
This phenomenon occurs when model parameters associated with previously learned tasks are rapidly overwritten by those for new tasks, resulting in a significant drop in performance on earlier tasks. While a complete retraining on all encountered data could theoretically address this issue, such time-consuming and resource-intensive solution is infeasible for real-world deployment. These fundamental challenges present a significant barrier to developing autonomous robotic agents capable of lifelong knowledge accumulation and complex long-horizon tasks adaptation in dynamic environments.

\ac{CL} offers a promising framework for developing versatile robotic agents capable of lifelong adaptation. Current methodologies primarily encompass five paradigms \cite{wang2024comprehensive}: regularization, experience replay, optimization, representation and parameter isolation. Regularization techniques impose constraints on weight updates to preserve critical parameters for previous tasks while learning new tasks \cite{kirkpatrick2017overcoming}. Replay-based approaches retain past knowledge through storing some samples \cite{rolnick2019experience} or generating pseudo-samples of previously learned tasks \cite{shin2017continual}. Optimization-based frameworks manipulate the optimization process through gradient projection techniques \cite{lopez2017gradient}. Representation-based methods leverage self-supervised learning \cite{pham2021dualnet} and large-scale pretraining \cite{wu2022class} to facilitate \ac{CL}. Parameter isolation approaches allocate dedicated network components for each task \cite{aljundi2017expert} to prevent any possible forgetting. 
These CL methods successfully retain knowledge from previously learned tasks and mitigate catastrophic forgetting while learning new tasks. However, these methods remain predominantly validated on image classification benchmarks \cite{lesort2020continual}, where maintaining recognition capabilities is sufficient. 
In contrast, robot manipulation learning requires concurrently incremental expansion of both perceptual and motor competencies \cite{thrun1995lifelong}. Several recent works \cite{roy2024m2distill, dong2025optimizing, li2023continual} have adapted these \ac{CL} algorithms to robotics with promising results. But they overlook the intrinsic features of robot manipulation tasks, leading to suboptimal performance particularly in complex long-horizon scenarios.  Therefore, an efficient and manipulation-centric CL algorithm for robotics remains an open challenge, necessitating solutions that balance generality with task-specific adaptability. 

\begin{figure*}[htbp]
\centering
\includegraphics[width=\textwidth]{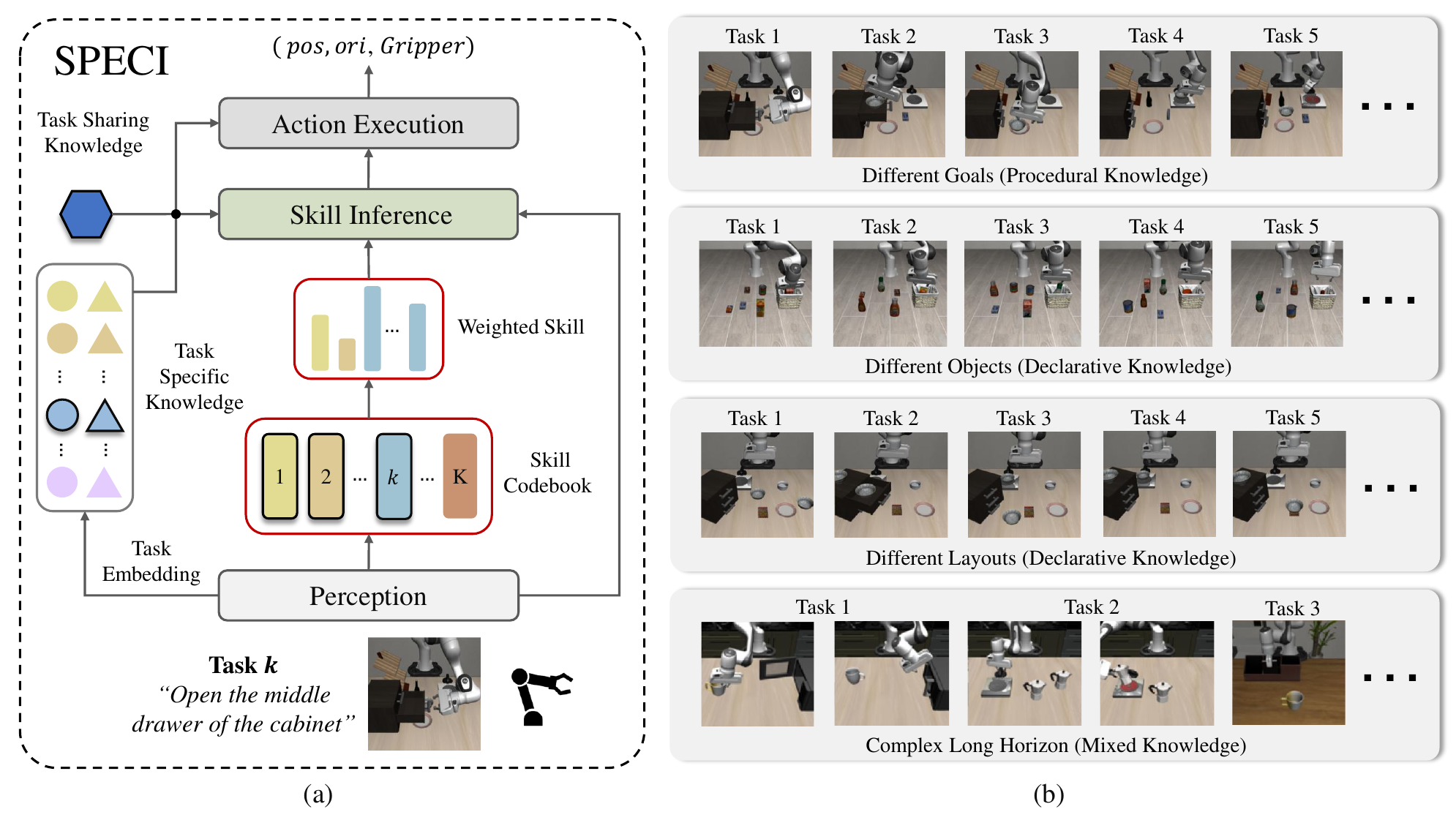}
\caption{(a) Overview of \ac{SPECI}, which consists of three hierarchical modules: Multimodal Perception, Skill Inference and Action Execution. (b) Illustration of four CL scenarios, each demanding distinct knowledge transfer capabilities. Goal states are visualized for all tasks except the \textit{Different Layouts} task suite, where initial states are displayed instead.}
\label{fig1}
\end{figure*}

To address the above challenges, we propose \ac{SPECI} (\underline{\textbf{S}}kill \underline{\textbf{P}}rompts-based Hi\underline{\textbf{E}}rarchical \underline{\textbf{C}}ontinual \underline{\textbf{I}}mitation Learning), an end-to-end hierarchical lifelong learning framework for robot manipulation. As illustrated in Fig. \ref{fig1}(a), the \ac{SPECI} architecture comprises three tightly-coupled components: 1) The Multimodal Perception and Fusion Module employs modality-specific encoders with subsequent cross-modal integration to process heterogeneous sensory inputs, producing unified state representations that maintain cross-modal task relevance; 2) The High-level Skill Inference Module dynamically retrieves skill embeddings from an extensible codebook based on contextual state representations and integrates them into the encoded observations; 3) The Low-level Action Execution Module generates parameterized action distributions to capture behavioral diversity present in skill executions. 
Differs from conventional flat policy architectures, our hierarchical framework enables context-aware skill selection and significantly improves performance in complex long-horizon tasks. Specifically, the skill inference module incorporates a dynamic skill codebook and an attention-driven skill selection mechanism. This innovative design empowers \ac{SPECI} to retain and reuse previously acquired skills while effectively learning new ones, achieving a good balance between stability and plasticity. Additionally, the codebook is learned autonomously alongside the policy, allowing \ac{SPECI} to interpret demonstrations independently and capture diverse, reasonable, and flexible skill representations. This approach eliminates the need for manual skill definitions and abstractions. The unified training paradigm facilitates the simultaneous learning of skills and their sequencing, resulting in efficient skill abstraction and smooth transitions between skills. To further enhance lifelong robot learning, we also incorporate task-specific and task-sharing knowledge into our model through mode approximation. Extensive experiments validate bidirectional knowledge transfer capabilities: prior skills improve the learning of new tasks (forward transfer), while new learned experiences refine the performance of previous tasks (backward transfer). 
The main contributions of this work can be summarized as follows:
\begin{enumerate}
\setlength{\itemindent}{3pt}  
    \item We develop \ac{SPECI}, a skill prompts based hierarchical \ac{CIL} framework that unifies multimodal perception and fusion, context-aware skill inference, and low-level action execution into a cohesive pipeline. This architecture significantly enhances the performance in complex long-horizon tasks.

    \item We propose an expansible skill codebook for implicit continual skill acquisition,  eliminating manual skill abstraction while promoting efficient skill-level knowledge transfer. Coupled with a decomposed attention-driven skill selection mechanism, our approach achieves  effective skill reutilization across diverse tasks.
    
    \item We introduce mode approximation to enrich our policy with task-specific and task-sharing knowledge through decomposing the attention-based parameters of a transformer decoder, thereby  enhancing knowledge transfer at the task level.

    \item Extensive experiments and ablation studies validate the effectiveness of our method, demonstrating that \ac{SPECI} achieves state-of-the-art performance in bidirectional knowledge transfer compared to existing approaches.
\end{enumerate}

\section{Related Work}
\subsection{Skill-based Imitation learning}
The challenge of policy learning has long been recognized both empirically and theoretically as scaling with the length of the problem horizon \cite{xie2021bellman}, which results in suboptimal performance in imitation learning for complex, long-horizon tasks. To mitigate this limitation, hierarchical decomposition of trajectories into reusable sensorimotor skills has emerged as a promising paradigm. By leveraging temporal abstraction, this approach effectively constrains the decision space and reduces the effective planning horizon in complex tasks \cite{zhao2023learning}. Such skill representations, alternatively referred to as subgoals \cite{st2024single} or primitives \cite{wang2023task}, can be derived either from human demonstrations \cite{sharma2019dynamics} or learned directly from raw sensory cues \cite{chu2019real}. Recent advances have demonstrated distinct approaches for skill acquisition. For instance, BUDS \cite{zhu2022bottom} adopts a bottom-up skill discovery paradigm,  autonomously extracting and organizing skills from unsegmented long-horizon demonstrations. Meanwhile, ACT \cite{zhao2023learning} employs autoregressive modeling of action sequences conditioned on visual observations to predict the sequence of future actions. They eliminate the requirement for pre-segmented demonstrations and allow robotic agents to handle complex, long-horizon manipulation tasks more efficiently. However, BUDS needs prior skill dataset partitioning and task-specific meta controllers to compose the skills for individual tasks, incurring substantial engineering overhead. Moreover, both methods follow a two-stage paradigm, where the benefits of temporal abstraction emerge only during the application phase after initial skill acquisition. Approaches such as DDCO \cite{krishnan2017ddco} jointly learn skills and action policies for a given task, but remain constrained by fixed state-action distributions. A critical limitation of existing methods is their reliance on a fixed set of skills in multitask settings, which becomes particularly problematic in \ac{CIL} scenarios where robotic agents must continually acquire new skills while retaining previously learned competencies. 
Our proposed \ac{SPECI} framework addresses these limitations through three key innovations: simultaneous abstraction of skill semantics and composition rules from cross-task demonstrations; dynamic codebook expansion via implicit skill prompts; an attention-driven skill adaptive selection mechanism that dynamically adjusts to changing environments and task goals. By incorporating skill retrieval and lifelong learning, \ac{SPECI} facilitates continual skill acquisition without the need for manual definition, overcoming the rigidity of prior approaches. 

\subsection{Continual Imitation Learning}
Continual imitation learning (CIL) aims to develop versatile agents capable of adapting to dynamic environments while preserving previously acquired competencies. 
The approach proposed in \cite{dong2025optimizing} employs partial data replay from previous tasks. Although this approach enhances knowledge retention, its reliance on explicit data storage poses inherent privacy risks in sensitive domains, while exhibiting catastrophic forgetting rates that scale proportionally with replay ratios. In contrast, generative methods such as CRIL \cite{gao2021cril} adopts deep generative replay (DGR) \cite{shin2017continual} to synthesize pseudo-demonstrations through action-conditioned video prediction. Although theoretically advantageous for privacy preservation, current implementations suffer from: insufficient capability in processing multimodal demonstration data, and unreliable video generation quality when handling long-horizon tasks with complex state transitions. Knowledge distillation techniques like PolyTask \cite{haldar2023polytask} attempt to unify policies through multitask learning, but their decoupled learning-distillation architecture inherently limits new task integration and bidirectional knowledge transfer. 
Recent advances, such as M2Distill \cite{roy2024m2distill}, address this issue through multi-modal distillation but remain constrained by their dependence on historical task data repositories, which are resource-intensive and not robust to noise. TAIL \cite{liu2023tail} leverages Low-Rank Adaptation (LoRA) to fine-tune large pretrained models for robot manipulation tasks. However, this approach demonstrates concerning plasticity degradation as task quantities increase, revealing fundamental scalability constraints. The above mentioned methods fail to adequately consider the intrinsic characteristics of robot manipulation, leading to inefficient and incomplete knowledge transfer. Our SPECI framework addresses these limitations by integrating both skill and task knowledge transfer through a hierarchical policy architecture that explicitly decomposes robotic motion into instructive skills and concrete actions.

\subsection{Skill-based Continual Imitation Learning}
The robotics community has reached a growing consensus that both declarative knowledge (entities and concepts) and procedural knowledge (task execution methodologies) play equally critical roles in continual robot learning \cite{liu2024libero}. This insight has driven the development of skill-based paradigms to improve knowledge transfer in \ac{CIL}. Some methods maintain expandable skill libraries to accommodate a growing number of manipulation tasks, frequently leveraging large language models \cite{tziafas2024lifelong}, \cite{parakh2024lifelong}. 
However, they are either limited to language modalities and stationary domains or they rely heavily on extensive user interactions, which reduces the system's autonomy and task execution efficiency.  
IsCiL \cite{lee2024incremental} addresses these challenges through an adapter-based skill learning architecture, utilizing a combination of multifaceted skill prototypes and an adaptive pool to model skill distributions for continual task adaptation. While effective, this approach needs sub-goal oriented skill retrieval during both training and evaluation, thereby introducing additional computational complexity and increasing resource demands. 
In contrast, LOTUS \cite{wan2024lotus} focuses on enabling robots to continuously learn and adapt to novel tasks through skill discovery and integration. This method employs an open-vocabulary vision model to identify skills and a meta-controller to integrate them effectively. Despite these advances, current methods still struggle with suboptimal bidirectional (backward and forward) knowledge transfer in vision-based manipulation scenarios.  
Our proposed \ac{SPECI} framework overcomes these limitations by implementing dynamic skill codebook expansion, attention-driven skill selection, and synchronized  acquisition and preservation of both task-specific and shared knowledge, collectively enabling superior bidirectional knowledge transfer capabilities.

\section{Proposed Method}
\subsection{Problem Formulation \& Overview of Our \ac{SPECI}}

This work focuses on developing a continual imitation learning approach to endow robotic agents with lifelong learning and adaptation capabilities, enabling persistent knowledge retention from previously learned manipulation tasks while dynamically accommodating novel objectives, scenes and tasks.

\textit{Robot Manipulation Tasks:} We formulate robot manipulation tasks as finite-horizon Markov Decision Processes $\mathcal{M} = (\mathcal{S},\mathcal{A},\mathcal{T},H,\mu_0,R)$, where $\mathcal{S}$ and $\mathcal{A}$ represent the state and action spaces, respectively, while $\mathcal{T}$, $H$, $\mu_0$, and $R$ denote the transition function, maximum episode horizon, initial state and reward, respectively. The state space $\mathcal{S}$ includes the RGB visual observations acquired from the wrist and workspace cameras, as well as joint angles and gripper states derived from proprioceptive feedback. The action space $\mathcal{A}$ encompasses target position, orientation, and gripper commands. Following \cite{liu2024libero}, sparse rewards is adopted with a goal predicate $g : \mathcal{S} \mapsto \{0, 1\}$ as $R$. The objective is to learn an optimal policy $\pi$ with maximum expected return:
\begin{equation}
\mathop{\max}_{\pi}J(\pi) = \mathop{\max}_{\pi} \mathbb{E}_{\boldsymbol{s_t}, \boldsymbol{a_t} \sim \pi, \mu_0} \left[ \sum_{t=1}^{H} g(\boldsymbol{s_t}) \right].
\end{equation}

\textit{Continual Imitation Learning:} We consider $K$ sequentially arriving manipulation tasks $\{T^1,\cdots, T^{K}\}$ to be learned by a unified policy $\pi(\cdot|\boldsymbol{s};\boldsymbol{l})$, conditioned on raw states $\boldsymbol{s}$ and language descriptors $\boldsymbol{l}$. Each task $T^k \equiv (\mu_0^k, g^k)$ shares the same $\mathcal{S},\mathcal{A}$, $\mathcal{T}$, and $H$, but has unique initial state $\mu_0^k$ and goal predicate $g^k$. When learning task $T^k$, the agent receives $N$ demonstrations $D^k = \left\{ \tau_n^k \right\}_{n=1}^{N}$, where $\tau_n^{k}=\left\{ (\boldsymbol{s^k_t}, \boldsymbol{a^k_t}) \right\}_{t=1}^{i_k}$, with episode horizon $i_k \leq H$. Adhering to a strict lifelong learning protocol that prevents access to previous tasks $\{D^j:j<k\}$ while training on $T^k$, we optimize the Behavioral Cloning (BC) \cite{bain1995framework} objective:
\begin{align}
& \mathop{\min}_{\pi}J_{\text{BC}}(\pi) = \\
& \mathop{\min}_{\pi} \frac{1}{k} \sum_{j=1}^{k} \mathbb{E}_{\boldsymbol{s^j_t}, \boldsymbol{a^j_t} \sim D^j} \left[ \sum_{t=0}^{i_j} \mathcal{L} \left( \pi(\boldsymbol{s^j_t}; \boldsymbol{l^j}), \boldsymbol{a^j_t} \right) \right] \nonumber,
\end{align}
where $\mathcal{L}$ denotes the negative log-likelihood loss over actions.

\begin{figure*}[htbp]
\centering
\includegraphics[width=\textwidth]{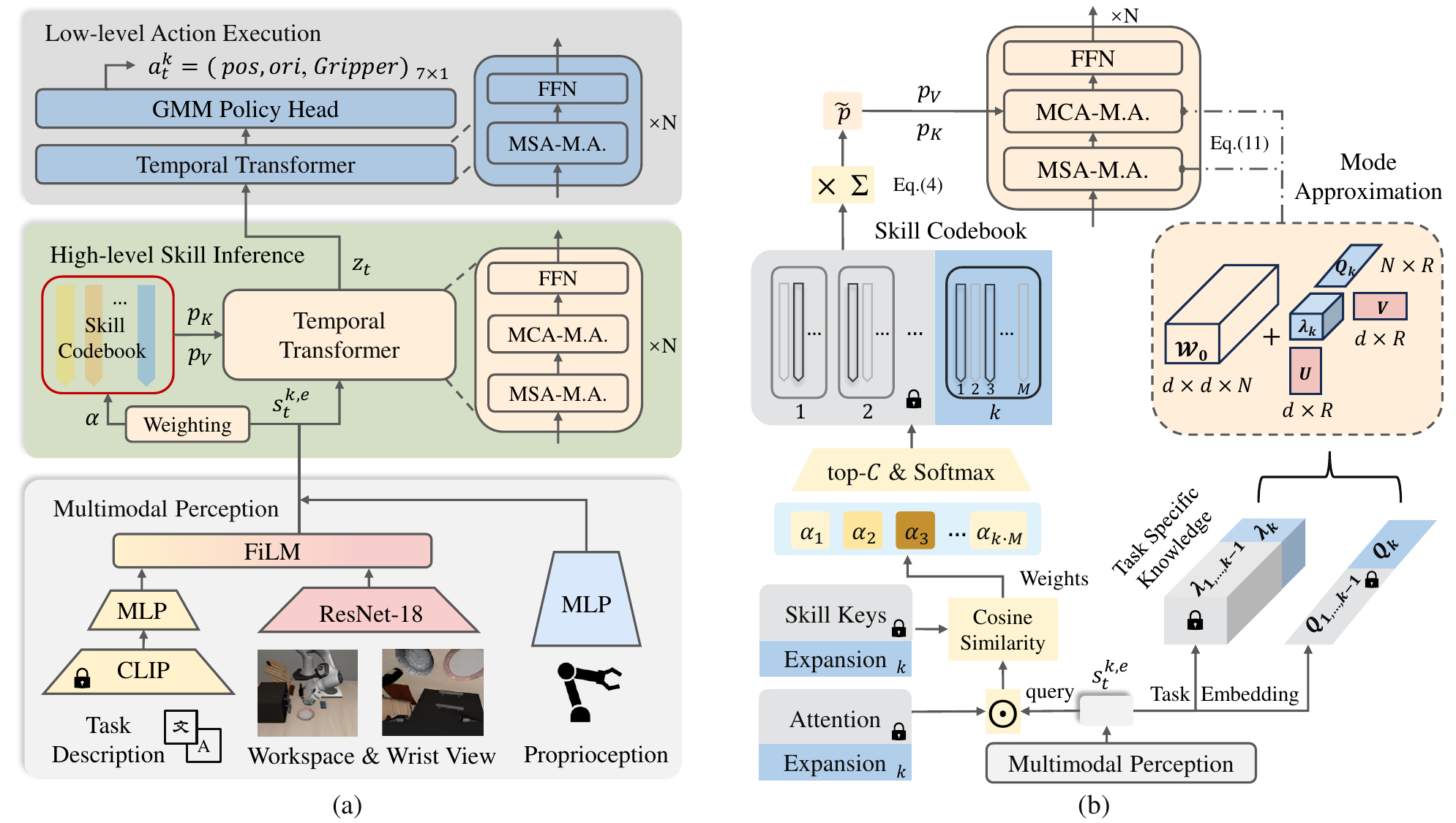}
\caption{(a) Framework of the proposed \ac{SPECI} for robot continual imitation learning. The \ac{SPECI} architecture consists of three hierarchical levels. The multimodal perception and fusion module encodes task descriptions, images of workspace and wrist views, and the robot’s proprioceptive state. In the high-level skill inference module, skill vectors are dynamically selected and weighted based on the output state embedding $\boldsymbol{s^{k,e}_t}$. The low-level action execution module then decodes and samples action $\boldsymbol{a^k_t}$ conditioned on the latent skill variable $\boldsymbol{z_t}$. Additionally, we enhance the transformer decoder in the last two levels with Mode Approximation to improve knowledge sharing and isolation across tasks. 
(b) Illustration of the skill selection and mode approximation mechanism. For the $k$-th task, \ac{SPECI} first initializes a new skill subset for the current task, along with the corresponding skill keys, attention vectors and task-specific decomposition vector $\boldsymbol{Q^k}$. Then, the synthesized latent skill $\boldsymbol{\Tilde{p}}$ is computed via a weighted summation according to the state information and is subsequently divided into $\boldsymbol{p_K}$ and $\boldsymbol{p_V}$. Finally, these components together with $\boldsymbol{Q^k}$ are input into the transformer decoder blocks to enhance knowledge transfer on both skill and task levels.}
\label{fig2}
\end{figure*}

\textit{Overview of Our \ac{SPECI}:} Continual learning methods from vision and NLP domains often fall short for robot manipulation due to their neglect of procedural knowledge and multimodal skill dependencies. Meanwhile, traditional flat policies in robot learning suffer from ineffective knowledge transfer and ambiguous skill abstraction. To overcome these limitations, we analyze the unique characteristics of robot learning and propose \ac{SPECI}, an end-to-end hierarchical \ac{CIL} architecture featuring dynamic skill codebook and mode approximation for robot manipulation, as shown in Fig. \ref{fig2}. \ac{SPECI} orchestrates perception, skill reasoning, and action generation through a three-level hierarchy: 1) multimodal perception level: a set of encoders processes and fuses multimodal sensory inputs; 2) high-level skill inference: a module dedicated to implicit skill abstraction and inference; 3) low-level action execution: a probabilistic policy that generates actions conditioned on skill contexts. This hierarchical decomposition, expressed as $\pi = \pi^H \cdot \pi^L$, avoids naively mimicking the expert's motion by separating high-level skill selection from low-level motor control through the latent skill variable $\boldsymbol{z_t}$:
\begin{equation}
    \pi(\boldsymbol{a^k_t}, \boldsymbol{z_t} \mid \boldsymbol{s^k_t}, \boldsymbol{l^k}) =
     \pi^H(\boldsymbol{z_t} \mid \boldsymbol{s^k_t}, \boldsymbol{l^k}) \pi^L(\boldsymbol{a^k_t} \mid \boldsymbol{z_t}, \boldsymbol{s^k_t}, \boldsymbol{l^k})
\end{equation}
In the context of \ac{CIL}, our hierarchical policy is trained on tasks presented sequentially, along with their corresponding demonstrations, in an end-to-end manner. The following subsections will successively introduce the three hierarchies of our approach and the mode approximation mechanism.

\subsection{Multimodal Perception and Fusion Module}
Effective robot manipulation fundamentally relies on multimodal perception, the synergistic integration of heterogeneous sensory streams that enables comprehensive environment understanding. While human beings naturally fuse visual, proprioceptive, and linguistic cues for precise manipulation, robotic agents require deliberate architectural design to achieve comparable sensorimotor coordination. To this end, we introduce the multimodal perception and fusion module as the first level of the \ac{SPECI} framework, as illustrated in Fig. \ref{fig2}(a). We leverage off-the-shelf networks to process raw sensory data from the robotic agent. For linguistic information, natural language descriptions of a task \eg \textit{``put the yellow and white mug in the microwave and close it"}, are input into CLIP \cite{radford2021learning}, a frozen pretrained text encoder to represent the task goal. The output is then converted into tokens using a \ac{MLP} layer. 
Regarding visual information, we utilize both wrist and workspace views. The wrist view captures the relative position between the robotic agent's end-effector and the object being manipulated, while the workspace view provides a comprehensive depiction of the overall situation, helping the robotic agent be aware of its state. 
These observation inputs are encoded with a pretrained ResNet-18 encoder \cite{he2016deep}. The visual encoder is updated alongside the policy learning process due to the significant gap between the robot manipulation scene and the database used for pretraining. Note that the choice of visual encoder is not limited to ResNet-18 and can be replaced with superior models in the future. Additionally, the language tokens are incorporated into the ResNet features using FiLM \cite{perez2018film}, allowing for the fusion of the two modalities. As a result, the extraction of image embeddings is modified and guided by current task goals, which is beneficial for accomplishing diverse goals within identical environments.
For proprioceptive state encoding, we train an \ac{MLP} from scratch to process joint angles and gripper status. 
At each timestep $t$, multimodal features are aggregated into a temporal state embedding $\boldsymbol{s^{k,e}_t} \in \mathbb{R}^{B \times L \times d}$, where $L$ is the sequence length multiplied by the number of modalities and $L=10 \times 5$ in our work. Then, the sequential states ${\{\boldsymbol{s^{k,e}_{t-4}},\cdots, \boldsymbol{s^{k,e}_{t}}, \cdots, \boldsymbol{s^{k,e}_{t+5}}\}}$ are passed to the following temporal transformer \cite{vaswani2017attention} to infer logical skills.

\subsection{High-level Skill Inference Module} 
Existing hierarchical methods depend on either manually crafted skills \cite{huang2023skill} or predetermined abstraction rules \cite{belkhale2024rt}, leading to an inherent trade-off in granularity. Finer skill granularity enhances the specificity of action guidance but complicates state-space learning, while coarser abstractions diminish executable precision. This dilemma often necessitates either extensive human annotations or computationally intensive perception modules \cite{belkhale2024rt}, ultimately limiting autonomous skill acquisition. In contrast, the proposed \ac{SPECI} addresses this granularity issue through unified skill-policy co-learning. Our \ac{SPECI} employs a dynamic skill codebook that automatically adapts abstraction levels according to task requirements while preserving  execution precision. This self-organizing mechanism enables task-contextualized skill extraction without the need for manual specification, optimizing both high-level planning and low-level control. 
Specifically, we decode the latent skill variable $\boldsymbol{z_t}$ using a temporal transformer with the input state embedding $\boldsymbol{s^{k,e}_t}$. Inspired by the prompt learning paradigm \cite{smith2023coda}, we introduce a weighted summation mechanism into our policy, as illustrated in Fig. \ref{fig2}(b). The synthesized latent skill is then divided and integrated into the key-value pairs of the temporal transformer to generate latent skill variables.

To enable consecutive skill acquisition and reuse, our skill codebook expands linearly with each new task rather than overwriting previously acquired skills. Our approach freezes existing skill subsets and optimizes only newly created skill vectors, effectively mitigating catastrophic forgetting at the skill level. Specifically, each task is allocated $M$ skill vectors, where $M$ is a hyperparameter controlling the the size of each skill subset. When learning the $k$-th task, the preceding $m = (k-1) \cdot M$ skill vectors remain locked while the newly initialized $k$-th skill subset captures the novel skills associated with the new task.  Additionally, there are some universal skills that appear across almost all tasks, such as \textit{move towards different directions along the coordinate axis} and \textit{close the gripper to grasp}. Our \ac{SPECI} inherently reuses these established skill vectors through not excluding them when formulating the current synthesized latent skill during training on new tasks or testing on previous tasks. This improves proficiency in tasks that were previously learned, as well as speeding up the learning procedure of new tasks. In addition, the robotic agent usually executes more than one skills simultaneously \cite{belkhale2024rt}. Therefore, instead of picking and choosing a single skill vector from the skill codebook, we select top-$C$ relevant skill vectors through attention-driven mechanism and combine them via weighted summation to formulate a synthesized latent skill. This approach allows the synthesized latent skill to capture an accurate combination of skills demanded by current state, resulting in more flexible and smooth motion for robotic agents. In general, the synthesized latent skill is computed by a weighted summation over the selected $C$ skill vectors:
\begin{equation}
    \boldsymbol{\Tilde{p}} = \sum_C \boldsymbol{\alpha_c} \boldsymbol{p_c},
\label{eq4}
\end{equation}
where $\boldsymbol{\alpha_c}$ denotes the weighting coefficients that modulate contributions of the corresponding skill vectors $\boldsymbol{p_c} \in \mathbb{R}^{B \times 2 \times d}$, with $d$ representing the embedding dimension. At each time step, the state embedding $\boldsymbol{s^{e}_t} \in \mathbb{R}^{B \times L \times d}$,  encoded by the multimodal perception and fusion module, serves as a query to create an attended query. This attended query is then utilized to compute cosine similarity $\gamma$ with the keys, yielding the weights of the current time step:
\begin{equation}\begin{aligned}
    \hspace{-1cm} 
    \boldsymbol{\alpha} & = \gamma(\boldsymbol{s_t^e} \odot \boldsymbol{A}, \boldsymbol{K}) \\
    & = [\{]\gamma(\boldsymbol{s_t^e} \odot \boldsymbol{A_1}, \boldsymbol{K_1}), \ldots, \gamma(\boldsymbol{s_t^e} \odot \boldsymbol{A_m}, \boldsymbol{K_m})],
\end{aligned}\end{equation}
where $\boldsymbol{K} \in \mathbb{R}^{B \times m \times d}$ denotes the keys corresponding to $m$ skill vectors, $\boldsymbol{A} \in \mathbb{R}^{B \times m \times d}$ is the associated attention vector of the skill codebook $\boldsymbol{P} \in \mathbb{R}^{B \times m \times 2 \times d}$ and $\odot$ indicates the Hadamard product. Unlike \cite{smith2023coda}, we found that imposing orthogonality constraints on $\boldsymbol{P}$, $\boldsymbol{K}$ and $\boldsymbol{A}$ is unnecessary, as applying Schmidt orthogonalization before learning new tasks is enough to prevent interference between key and skill vectors acquired across different tasks. Then, the top-$C$ relevant skill weights $\alpha_c$ can be computed as: 
\begin{equation}
    \boldsymbol{\alpha_c} = Softmax(\operatorname{Top-C}(\boldsymbol{\alpha})).
\end{equation}
The synthesized latent skill $\boldsymbol{\tilde{p}}$ obtained from Eq. (\ref{eq4}) undergoes dimensional factorization to generate parameter pairs ${\boldsymbol{p_K}, \boldsymbol{p_V}} \in \mathbb{R}^{B \times 1 \times d}$, which are subsequently injected into all \ac{MCA} layers via prefix-tuning (P-T):
\begin{equation}
   f_{P-T}(\boldsymbol{\Tilde{p}}, \boldsymbol{s_t^e}) = \text{MCA}(\boldsymbol{h_q}, [\boldsymbol{p_K}; \boldsymbol{h_k}], [\boldsymbol{p_V}; \boldsymbol{h_v}]).
\end{equation}
where $\boldsymbol{h_q}$, $\boldsymbol{h_k}$ and $\boldsymbol{h_v}$ denote the query, key and value projections. 
The latent skill variable $\boldsymbol{z_t}$ is generated through a series of $N$ transformer decoder blocks. In both the \ac{MSA} and \ac{MCA} layers, the query $\boldsymbol{h_q}$, key $\boldsymbol{h_k}$ and value $\boldsymbol{h_v}$ share the same input embedding $\boldsymbol{s_t^e}$. 

\subsection{Low-level Action Execution Module}
The temporal sequence of latent skill variables $\{\boldsymbol{z_{t-4}},\cdots, \boldsymbol{z_{t}},\cdots,\boldsymbol{z_{t+5}}\}$ is processed through another temporal transformer decoder to generate executable actions. Unlike simple action sequence regression, which fails to capture the rich multimodal distribution of expert demonstrations \cite{wang2023mimicplay}, we employ an \ac{MLP}-based \ac{GMM} policy head to sample actions, following recent approaches in \cite{liu2024libero} and \cite{wan2024lotus}. The \ac{GMM} formulation in our action execution module is expressed as: 
\begin{equation}
    p(\boldsymbol{\tau} \mid \boldsymbol{\theta}) = \sum_{\boldsymbol{\gamma}} p(\boldsymbol{\tau} \mid \boldsymbol{\theta}, \boldsymbol{\gamma}) p(\boldsymbol{\gamma} \mid \boldsymbol{\theta}),
\end{equation}
where $\boldsymbol{\theta} = \{\boldsymbol{\mu_r}, \boldsymbol{\sigma_r}, \eta_r\}_{r=1}^{R}$ parameterizes the \ac{GMM} policy head. The action trajectory $\boldsymbol{\tau}$'s distribution is modeled using $R$ Gaussian components, where each component is given by $p(\boldsymbol{\tau} \mid \boldsymbol{\theta}, \boldsymbol{\gamma_r}) = \mathcal{N}(\boldsymbol{\tau} \mid \boldsymbol{\mu_r}, \boldsymbol{\sigma_r})$, with $\boldsymbol{\gamma_r}$ indicating the parameters of the $r$-th gaussian component. $\boldsymbol{\gamma_r}$ is comprised of mean $\boldsymbol{\mu_r}$, standard deviation $\boldsymbol{\sigma_r}$, and corresponding weight $\eta_r$.
Consequently, the learning objective for our \ac{GMM} model \ie the \ac{SPECI} loss for learning the $k$-th task, is to minimize the negative log-likelihood over all the corresponding action trajectories $\boldsymbol{\tau^k}$ of the expert demonstrations: 
\begin{equation}
    \mathcal{L}_{\text{GMM}}(\boldsymbol{\theta}) = -\mathbb{E}_{\boldsymbol{\tau^k}} \log \left( \sum_{r=1}^{R} \eta_r \mathcal{N}(\boldsymbol{\tau^k} \mid \boldsymbol{\mu_r}, \boldsymbol{\sigma_r}) \right),
\end{equation}
where $0 \leq \eta_r \leq 1$ and $ \sum_{r=1}^{R} \eta_r = 1$.

\subsection{Mode Approximation}

In lifelong robot learning, effectively leveraging both task-specific and task-sharing knowledge is crucial for achieving higher efficiency and satisfying adaptability \cite{lee2024incremental}. While the dynamic skill codebook successfully enables procedural knowledge transfer, we simultaneously enhance declarative knowledge transfer at task level through parameter-efficient adaptation. 
Building upon Parameter-Efficient Fine-Tuning (PEFT) principles \cite{he2021towards}, we update the declarative knowledge with a few additional trainable parameters and introduce mode approximation to our work to isolate task-specific knowledge while facilitating task-analogous knowledge sharing, as illustrated in Fig. \ref{fig2}(b). Inspired by \cite{wang2023parameter}, we regard the attention-based parameters of a transformer block as a 3D tensor $\boldsymbol{\mathcal{W}^o} \in \mathbb{R}^{d \times d \times N}$, where $N$ denotes the number of weight matrices. 
Through tensor decomposition, we extract low-rank yet discriminative non-redundant task features $\boldsymbol{W_k}$ for each task effectively, which are subsequently used to augment $\boldsymbol{\mathcal{W}^o}$ efficiently.

In conventional tensor decomposition, a 3D tensor comprising three modes can be projected along each respective axis to reduce its dimensionality. 
Among existing approaches, \ac{CP} decomposition \cite{tucker1966some} stands out for its lightweight nature and interpretability. Inspired by this, we adopt \ac{CP} decomposition to factorize the 3D trainable parameter tensor $\boldsymbol{W_k} \in \mathbb{R}^{d \times d \times N}$ into a sum of $R$ rank-one components, effectively capturing the intrinsic task-specific features. Each component is expressed as the outer product of three decomposed vectors  \cite{wang2023parameter}:   
\begin{equation}
    \boldsymbol{W_k} = \sum_{r=1}^{R} \lambda_r (\boldsymbol{u_r} \circ \boldsymbol{v_r} \circ \boldsymbol{q_r}),
\end{equation}
where $\boldsymbol{u_r} \in \mathbb{R}^d$, $\boldsymbol{v_r} \in \mathbb{R}^d$ and $\boldsymbol{q_r} \in \mathbb{R}^N$ are the decomposed factors. $\lambda_r$ is the scaling coefficient for the $r$-th component, and $R$ denotes the rank of the \ac{CP} decomposition. 

In particular, $\boldsymbol{U}, \boldsymbol{V}$ serve as global factors that capture shared knowledge across tasks, while $\boldsymbol{Q^k}$ and the learnable coefficient vector $\boldsymbol{\lambda^k} \in \mathbb{R}^R$ (randomly initialized) are in charge of capturing the discriminative features of each task. We directly associate task embeddings to $\boldsymbol{Q^k}$ and $\boldsymbol{\lambda^k}$. The mode factors $\boldsymbol{U} = \begin{bmatrix} \boldsymbol{u_1}, \cdots, \boldsymbol{u_R} \end{bmatrix}$, $\boldsymbol{Q^k} = \begin{bmatrix} \boldsymbol{q^k_1}, \cdots, \boldsymbol{q^k_R} \end{bmatrix}$ are randomly initialized with Gaussian distribution, while $\boldsymbol{V} = \begin{bmatrix} \boldsymbol{v_1}, \cdots, \boldsymbol{v_R} \end{bmatrix}$ is initialized with zeros in preparation for the approximation of the attention-based weight in all the transformer blocks. Through these three mode factors, the mode approximation in the forward propagation can be achieved by the inverse progress of \ac{CP} decomposition as follows:
\begin{equation}
    \boldsymbol{\mathcal{H}^k} = \boldsymbol{\mathcal{W}^o} \boldsymbol{\mathcal{X}^k} + \left( \sum_{r=1}^{R} \lambda_r^k (\boldsymbol{u_r} \circ \boldsymbol{v_r} \circ \boldsymbol{q_r}) \right) \boldsymbol{\mathcal{X}^k},
\end{equation}
where $\boldsymbol{\mathcal{H}^k}$ represents the hidden output of either the \ac{MSA} or \ac{MCA} module, and $\boldsymbol{\mathcal{X}_k}$ denotes the input tensor, corresponding to either the state embedding $\boldsymbol{s_t^{k,e}}$ or the latent skill variable $\boldsymbol{z_t}$ in our work. 

\section{Experiments}

This section presents a comprehensive experimental evaluation of \ac{SPECI} across two key dimensions: comparative analysis of forward and backward transfer performance against state-of-the-art methods, and assessment of continual learning capabilities relative to the upper bounds. Furthermore, we conduct systematic ablation studies to quantitatively evaluate the individual contributions of three core components: the hierarchical policy architecture, the dynamic skill codebook, and the mode approximation mechanism. These investigations provide empirical evidence for their respective roles in enhancing knowledge transfer within lifelong learning scenarios.

\subsection{Experimental Setup}
\subsubsection{Datasets}
We evaluate our method on the LIBERO dataset \cite{liu2024libero}, a benchmark specifically designed for lifelong robot learning that systematically assesses  knowledge transfer across sequential manipulation tasks. 
The benchmark encompasses a diverse set of tasks that exhibit substantial diversity in both task goals (ranging from basic manipulation skills such as \textit{pick\&place} and \textit{push\&pull} to complex sequential motions) and environmental contexts (covering multiple household settings including kitchens, living rooms, and study rooms), with detailed visualizations provided in Fig. \ref{fig1} (b). The dataset is divided into four task suites and each consists of 10 tasks:
\begin{enumerate}[label =(\alph*)]
    \item LIBERO-OBJECT: This suite focuses on the identification and placement of distinct objects, requiring a robot to continually learn about various object types (declarative knowledge). 
    \item LIBERO-GOAL: This suite challenges the robot to accomplish a range of task goals, allowing for the evaluation of its procedural knowledge transfer capabilities. 
    \item LIBERO-SPATIAL: This suite requires a robot to distinguish two identical bowls based solely on their spatial context and place one on a plate, assessing its ability to memorize and understand spatial relationships (declarative knowledge).
    \item LIBERO-LONG: This task suite involves long-horizon tasks featuring two sub-goals with different objects and scenes, testing the robot's ability for  synthesized continual learning on complex tasks. 
\end{enumerate}

\subsubsection{Metrics}
Following \cite{liu2024libero}, three standard metrics are employed to evaluate policy performance: \ac{FWT}, \ac{NBT}, and \ac{AUC}.
\ac{FWT} quantifies policy adaptation speed to new tasks, with higher \ac{FWT} values indicating more efficient learning and better prior knowledge integration. \ac{NBT} measures how much knowledge of previous tasks the policy forgets when learning new ones, where lower \ac{NBT} values denote stronger retention of performance on seen tasks. \ac{AUC} provides a holistic measure of average success rates across all tasks, with higher \ac{AUC} values implying superior overall performance considering both \ac{FWT} and \ac{NBT}. They are defined as:
\begin{align}
    & \text{FWT} = \sum_{k=1}^K\frac{\text{FWT}_k}{K } = \sum_{k=1}^K\frac{\sum_{e \in \{0 \ldots 50\}} c_{k,k,e}}{K \times 11}\\
    & \text{NBT} = \sum_{k=1}^K\frac{1}{K(K-k)} \sum_{q=k+1}^{K} (c_{k,k} - c_{q,k}) \\
    & \text{AUC} = \sum_{k=1}^K \frac{1}{K(K-k+1)} \left( \text{FWT}_k + \sum_{q=k+1}^{K} c_{q,k} \right)
\end{align}
where $K$ is the total number of tasks, \( c_{i,j,e} \) denotes the agent's success rate on task \( j \) after learning previous $i-1$ tasks and \( e \) epochs (\( e \in \{0, 5, \dots, 50\} \)) on task \( i \). Define \( c_{i,i} = \max_e c_{i,i,e} \) as the maximum success rate on task \( i \). Let \( e_i^* \) be the earliest epoch where this maximum is achieved, i.e., $e_i^* = \arg\min_e \{ c_{i,i,e} = c_{i,i} \}$. For all \( e \geq e_i^* \), assume \( c_{i,i,e} = c_{i,i} \). For \( j \neq i \), define \( c_{i,j} = c_{i,j,e_i^*} \).

\begin{table*}[ht]
    \centering
    \caption{Performance comparison of different baselines and policy architectures under the ER learning paradigm across all four LIBERO task suites. We report the mean and standard deviation of FWT, NBT and AUC values averaged over three random seeds, with the best results highlighted in bold.}
    \label{tab_policy_archi}
    \begin{tabular}{llcccccc}
        \toprule
        \textbf{Algorithm} & \textbf{Policy Architecture} & \textbf{FWT($\uparrow$)} & \textbf{NBT($\downarrow$)} & \textbf{AUC($\uparrow$)} & \textbf{FWT($\uparrow$)} & \textbf{NBT($\downarrow$)} & \textbf{AUC($\uparrow$)} \\
        \midrule
        \multicolumn{2}{c}{} & \multicolumn{3}{c}{\textbf{LIBERO-OBJECT}} & \multicolumn{3}{c}{\textbf{LIBERO-GOAL}} \\
        \cmidrule(lr){3-5} \cmidrule(lr){6-8}
        \multirow{1}{*}{\textsc{Sequential}} 
        & \ac{SPECI} (Ours) & $0.80 \pm 0.02$ & $0.68 \pm 0.02$ & $0.30 \pm 0.01$ & $0.81 \pm 0.00$ & $0.81 \pm 0.01$ & $0.23 \pm 0.00$ \\
        \cmidrule(lr){2-8}
        \multirow{6}{*}{\textsc{ER}}
        & \textsc{ResNet-RNN} & $0.30 \pm 0.01$ & $0.27 \pm 0.05$ & $0.17 \pm 0.05$ & $0.41 \pm 0.00$ & $0.35 \pm 0.01$ & $0.26 \pm 0.01$ \\
        & \textsc{ResNet-T} & $0.67 \pm 0.07$ & $0.43 \pm 0.04$ & $0.44 \pm 0.06$ & $0.64 \pm 0.01$ & $0.34 \pm 0.02$ & $0.49 \pm 0.02$ \\
        & \textsc{VIT-T} & $0.70 \pm 0.02$ & $0.28 \pm 0.01$ & $0.57 \pm 0.01$ & $0.57 \pm 0.00$ & $0.40 \pm 0.02$ & $0.38 \pm 0.01$ \\
        & \textsc{BUDS \cite{zhu2022bottom}} & $0.52 \pm 0.02$ & $0.21 \pm 0.01$ & $0.47 \pm 0.01$ & $0.50 \pm 0.01$ & $0.39 \pm 0.01$ & $0.42 \pm 0.01$ \\
        & \textsc{LOTUS \cite{wan2024lotus}} & $0.74 \pm 0.03$ & $0.11 \pm 0.01$ & $0.65 \pm 0.03$ & $0.61 \pm 0.03$ & $0.30 \pm 0.01$ & $0.56 \pm 0.01$ \\
        & \textbf{SPECI (Ours)} & \textbf{0.83} $\pm$ \textbf{0.01} & \textbf{0.10} $\pm$ \textbf{0.02} & \textbf{0.78} $\pm$ \textbf{0.03} & \textbf{0.74} $\pm$ \textbf{0.01} & \textbf{0.20} $\pm$ \textbf{0.02} & \textbf{0.65} $\pm$ \textbf{0.00} \\
        \cmidrule(lr){2-8}
        \multirow{1}{*}{\textsc{Multitask}}
        & \ac{SPECI} (Ours) & \multicolumn{3}{r}{$0.87 \pm 0.00$} & \multicolumn{3}{r}{$0.90 \pm 0.01$} \\
        \midrule
        \multicolumn{2}{c}{} & \multicolumn{3}{c}{\textbf{LIBERO-SPATIAL}} & \multicolumn{3}{c}{\textbf{LIBERO-LONG}} \\
        \cmidrule(lr){3-5} \cmidrule(lr){6-8}
        \multirow{1}{*}{\textsc{SeqL}} 
        & \ac{SPECI} (Ours) & $0.75 \pm 0.00$ & $0.68 \pm 0.01$ & $0.23 \pm 0.00$ & $0.57 \pm 0.00$ & $0.57 \pm 0.02$ & $0.18 \pm 0.00$ \\
        \cmidrule(lr){2-8}
        \multirow{4}{*}{\textsc{ER}} 
        & \textsc{ResNet-RNN} & $0.40 \pm 0.02$ & $0.29 \pm 0.02$ & $0.29 \pm 0.01$ & $0.16 \pm 0.02$ & \textbf{0.16} $\pm$ \textbf{0.02} & $0.08 \pm 0.01$ \\
        & \textsc{ResNet-T} & $0.65 \pm 0.03$ & $0.27 \pm 0.03$ & $0.56 \pm 0.01$ & $0.48 \pm 0.02$ & $0.32 \pm 0.04$ & $0.32 \pm 0.01$ \\
        & \textsc{VIT-T} & $0.63 \pm 0.01$ & $0.29 \pm 0.02$ & $0.50 \pm 0.02$ & $0.38 \pm 0.05$ & $0.29 \pm 0.06$ & $0.25 \pm 0.02$ \\
        & \textbf{SPECI (Ours)} & \textbf{0.67} $\pm$ \textbf{0.00} & \textbf{0.06} $\pm$ \textbf{0.01} & \textbf{0.66} $\pm$ \textbf{0.02} & \textbf{0.58} $\pm$ \textbf{0.01} & $0.21 \pm 0.01$ & \textbf{0.46} $\pm$ \textbf{0.00} \\
        \cmidrule(lr){2-8}
        \multirow{1}{*}{\textsc{Multitask}} 
        & \ac{SPECI} (Ours) & \multicolumn{3}{r}{$0.86 \pm 0.00$} & \multicolumn{3}{r}{$0.70 \pm 0.00$} \\
        \bottomrule
    \end{tabular}
\end{table*}

\subsubsection{Comparison Methods}
We evaluate \ac{SPECI} against the three benchmark architectures in \cite{liu2024libero}, and two state-of-the-art methods:  BUDS \cite{zhu2022bottom} and LOTUS \cite{wan2024lotus}. These comparison approaches encompass both skill-based hierarchical architectures and classical flat imitation learning structures:
\begin{enumerate}
    \item BENCHMARK ARCHITECTURES \cite{liu2024libero}: Three flat vision-language policy network architectures: RESNET-RNN, RESNET-T, and VIT-T, serving as fundamental comparisons.
    \item BUDS \cite{zhu2022bottom}: A hierarchical policy model that executes multitask skill discovery to develop its policies and adapts to new tasks through retraining at each lifelong learning step. 
    \item LOTUS \cite{wan2024lotus}: A hierarchical imitation learning framework with experience replay (ER) that employs an open-world vision model for continual skill discovery and extraction from unsegmented demonstrations. A meta-controller integrates these skills to manage robot manipulation tasks, enabling lifelong robot learning.
\end{enumerate}

The following baselines and representative continual learning techniques are employed in our experiments:
\begin{enumerate}
    \item SEQUENTIAL: A naive CL baseline that fine-tunes the network incrementally on each new task, which represents the upper bound for \ac{FWT} as it fully adapts to each new task at the cost of catastrophic forgetting.
    \item MULTITASK: An oracle baseline trained jointly on all tasks with full access to all task data. We report its average success rate as the upper bound of \ac{AUC}. 
    \item ER \cite{chaudhry2019tiny}: A replay-based CL method that maintains a fixed-size replay buffer  (1000 trajectories in our implementation)  storing exemplars from previous tasks to mitigate catastrophic forgetting. 
    \item PACKNET \cite{mallya2018packnet}: A parameter isolation approach that protects task-specific network parameters during CL by freezing important parameters from previous tasks while allocating new capacity for incoming tasks. 
\end{enumerate}

\subsubsection{Implementation Details}
All experiments are conducted on an NVIDIA RTX 4090 GPU, following the standardized training and evaluation protocols established in LIBERO \cite{liu2024libero}. We adjust the image embedding dimension to 384 and expand the transformer MLP hidden size to 1536. The model was trained for 50 epochs per lifelong learning step using a learning rate of 1e-4 with weight decay. During training, We implement an early stopping criterion that triggered when the success rate declined twice consecutively after achieving 95\% performance threshold. We empirically set the number of skill vectors $M$ and $C$ to be 10, and the rank $R$ of CP decomposition to be 64. Note that all the tasks share a single task-specific parameter $\textbf{Q}$ when implemented with ER \cite{chaudhry2019tiny}, SEQUENTIAL and MULTITASK baselines.

\begin{figure*}[htbp]
    \centering
    \subfloat{\label{subfig:er}\includegraphics[width=0.95\textwidth]{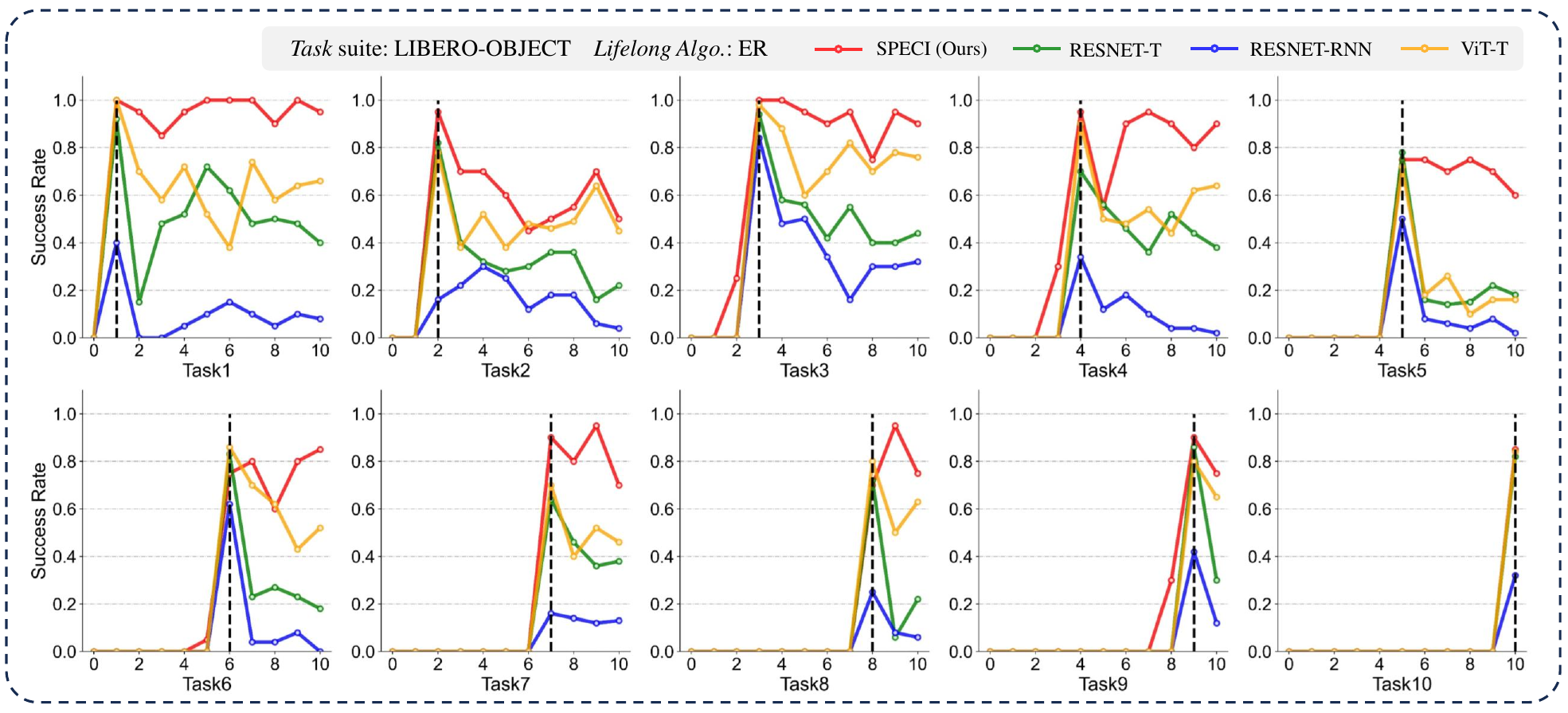}} \\
    \vspace{0mm} 
    \subfloat{\label{subfig:packnet}\includegraphics[width=0.95\textwidth]{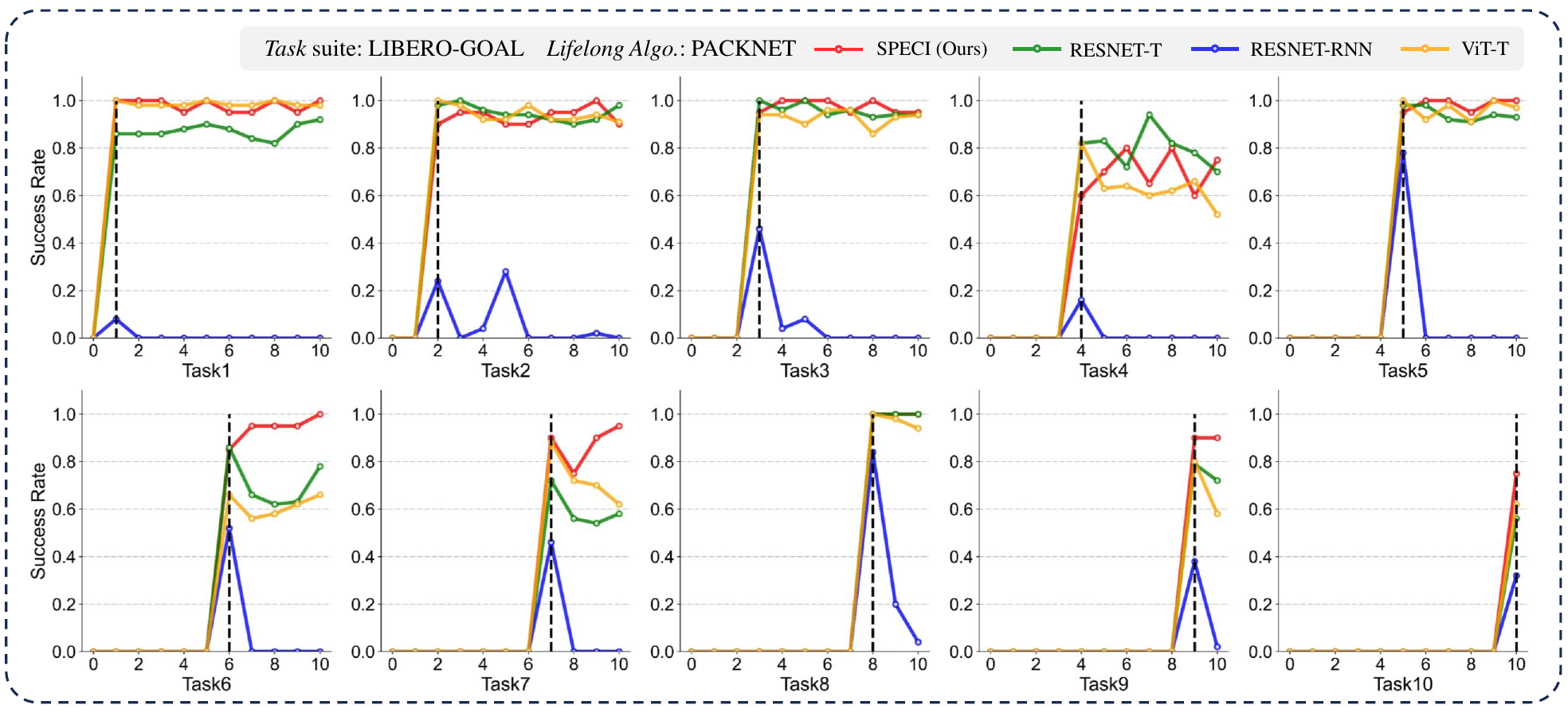}}
    \centering
    \caption{Comparison of different policy architectures under ER \cite{chaudhry2019tiny} and PACKNET \cite{mallya2018packnet} lifelong learning paradigms, evaluated on LIBERO-OBJECT and LIBERO-GOAL task suites. We report the success rate for each task throughout the whole continual learning procedure. Each \textit{Task i} subplot illustrates the agent's performance on the $i$-th task after training on the corresponding task ($x$-axis).}
    \label{er_packnet_each_task_suc}
\end{figure*}

\begin{table*}[htbp]
    \centering
    \caption{Performance comparison of different baselines and policy architectures under the PACKNET learning paradigm across all four LIBERO task suites. We report the mean and standard deviation of FWT, NBT and AUC values averaged over three random seeds, with the best results highlighted in bold.}
    \label{tab_lifelong_algor}
    \begin{tabular}{llcccccc}
        \toprule
        \textbf{Algorithm} & \textbf{Policy Architecture} & \textbf{FWT($\uparrow$)} & \textbf{NBT($\downarrow$)} & \textbf{AUC($\uparrow$)} & \textbf{FWT($\uparrow$)} & \textbf{NBT($\downarrow$)} & \textbf{AUC($\uparrow$)} \\
        \midrule
        \multicolumn{2}{c}{} & \multicolumn{3}{c}{\textbf{LIBERO-OBJECT}} & \multicolumn{3}{c}{\textbf{LIBERO-GOAL}} \\
        \cmidrule(lr){3-5} \cmidrule(lr){6-8}
        \multirow{1}{*}{\textsc{Sequential}} 
        & \ac{SPECI} (Ours) & $0.80 \pm 0.02$ & $0.68 \pm 0.02$ & $0.30 \pm 0.01$ & $0.81 \pm 0.00$ & $0.81 \pm 0.01$ & $0.23 \pm 0.00$ \\
        \cmidrule(lr){2-8}
        \multirow{4}{*}{\textsc{PackNet}}
        & \textsc{ResNet-RNN} & $0.29 \pm 0.02$ & $0.35 \pm 0.02$ & $0.13 \pm 0.01$ & $0.32 \pm 0.03$ & $0.37 \pm 0.04$ & $0.11 \pm 0.01$ \\
        & \textsc{ResNet-T} & $0.60 \pm 0.07$ & $0.17 \pm 0.05$ & $0.60 \pm 0.05$ & $0.63 \pm 0.02$ & $0.06 \pm 0.01$ & $0.75 \pm 0.01$ \\
        & \textsc{VIT-T} & $0.58 \pm 0.03$ & $0.18 \pm 0.02$ & $0.56 \pm 0.04$ & $0.69 \pm 0.02$ & $0.08 \pm 0.01$ & $0.76 \pm 0.02$ \\
        & \textbf{SPECI (Ours)} & \textbf{0.81} $\pm$ \textbf{0.03} & \textbf{-0.01} $\pm$ \textbf{0.02} & \textbf{0.85} $\pm$ \textbf{0.03} & \textbf{0.81} $\pm$ \textbf{0.01} & \textbf{-0.01} $\pm$ \textbf{0.00} & \textbf{0.87} $\pm$ \textbf{0.01} \\
        \cmidrule(lr){2-8}
        \multirow{1}{*}{\textsc{Multitask}}
        & \ac{SPECI} (Ours) & \multicolumn{3}{r}{$0.87 \pm 0.00$} & \multicolumn{3}{r}{$0.90 \pm 0.01$} \\
        \midrule
        \multicolumn{2}{c}{} & \multicolumn{3}{c}{\textbf{LIBERO-SPATIAL}} & \multicolumn{3}{c}{\textbf{LIBERO-LONG}} \\
        \cmidrule(lr){3-5} \cmidrule(lr){6-8}
        \multirow{1}{*}{\textsc{SeqL}} 
        & \ac{SPECI} (Ours) & $0.75 \pm 0.00$ & $0.68 \pm 0.01$ & $0.23 \pm 0.00$ & $0.57 \pm 0.00$ & $0.57 \pm 0.02$ & $0.18 \pm 0.00$ \\
        \cmidrule(lr){2-8}
        \multirow{4}{*}{\textsc{PackNet}} 
        & \textsc{ResNet-RNN} & $0.27 \pm 0.03$ & $0.38 \pm 0.03$ & $0.06 \pm 0.01$ & $0.13 \pm 0.00$ & $0.21 \pm 0.01$ & $0.03 \pm 0.00$ \\
        & \textsc{ResNet-T} & $0.55 \pm 0.01$ & $0.07 \pm 0.02$ & $0.63 \pm 0.00$ & $0.22 \pm 0.01$ & $0.08 \pm 0.01$ & $0.25 \pm 0.00$ \\
        & \textsc{VIT-T} & $0.57 \pm 0.04$ & $0.15 \pm 0.00$ & $0.59 \pm 0.03$ & $0.36 \pm 0.01$ & $0.14 \pm 0.01$ & $0.34 \pm 0.01$ \\
        & \textbf{SPECI (Ours)} & \textbf{0.76} $\pm$ \textbf{0.02} & \textbf{-0.04} $\pm$ \textbf{0.01} & \textbf{0.77} $\pm$ \textbf{0.00} & \textbf{0.61} $\pm$ \textbf{0.01} & \textbf{-0.01} $\pm$ \textbf{0.00} & \textbf{0.61} $\pm$ \textbf{0.01} \\
        \cmidrule(lr){2-8}
        \multirow{1}{*}{\textsc{Multitask}} 
        & \ac{SPECI} (Ours) & \multicolumn{3}{r}{$0.86 \pm 0.00$} & \multicolumn{3}{r}{$0.70 \pm 0.00$} \\
        \bottomrule
    \end{tabular}
\end{table*}

\subsection{Main Experimental Results}

Table \ref{tab_policy_archi} presents a comprehensive comparison of our \ac{SPECI} method against current state-of-the-art approaches under the ER lifelong learning paradigm. In general, skill-based hierarchical policies, such as BUDS \cite{zhu2022bottom}, LOTUS \cite{wan2024lotus}, and our \ac{SPECI}, consistently outperform naive flat benchmark policies \cite{liu2024libero} in \ac{NBT}, demonstrating the benefits of skill guidance in preserving both procedural and declarative knowledge. 
LOTUS \cite{wan2024lotus} outperforms BUDS \cite{zhu2022bottom} due to its incorporation of a transformer encoder and a well-designed binary mask for skill prediction. Our method achieves the lowest \ac{NBT} across three task suites, demonstrating robust backward knowledge transfer capability. In particular, \ac{SPECI} achieves a remarkable 21\% reduction in \ac{NBT} on LIBERO-SPATIAL compared to the best previous result, indicating effective skill preservation and reapplication. Although the RESNET-RNN policy exhibits lower \ac{NBT} on LIBERO-LONG, this apparent advantage arises from its fundamentally limited learning capacity, as evidenced by its poorer performance in both \ac{FWT} and \ac{AUC} metrics. The model's limited success rate naturally reduces the potential for catastrophic forgetting to occur, as there is less acquired knowledge to be forgotten in the first place. Our \ac{SPECI} achieves competitive performance, indicating its superiority in dealing with complex long-horizon tasks. In terms of \ac{FWT}, BUDS \cite{zhu2022bottom} fails to show meaningful improvement compared to flat policies, primarily due to its simplistic retraining approach that prevents effective utilization of previously acquired skills for new task acquisition. In contrast, LOTUS \cite{wan2024lotus} achieves better \ac{FWT} performance owing to its advanced semantic feature extractor and the \ac{GMM} policy head. Our \ac{SPECI} method outperforms state-of-the-art baselines by 9\%, 10\%, 2\%, and 10\% across all task suites, demonstrating strong forward knowledge transfer through prior skill reuse and task-sharing knowledge acquisition. In terms of \ac{AUC}, while BUDS \cite{zhu2022bottom} lags behind LOTUS \cite{wan2024lotus}, our \ac{SPECI} achieves consistent improvements of 13\%, 9\%, 10\%, and 14\% over competing methods, highlighting its superior overall performance.

\begin{figure}[ht]
\centering
\includegraphics[width=0.49\textwidth]{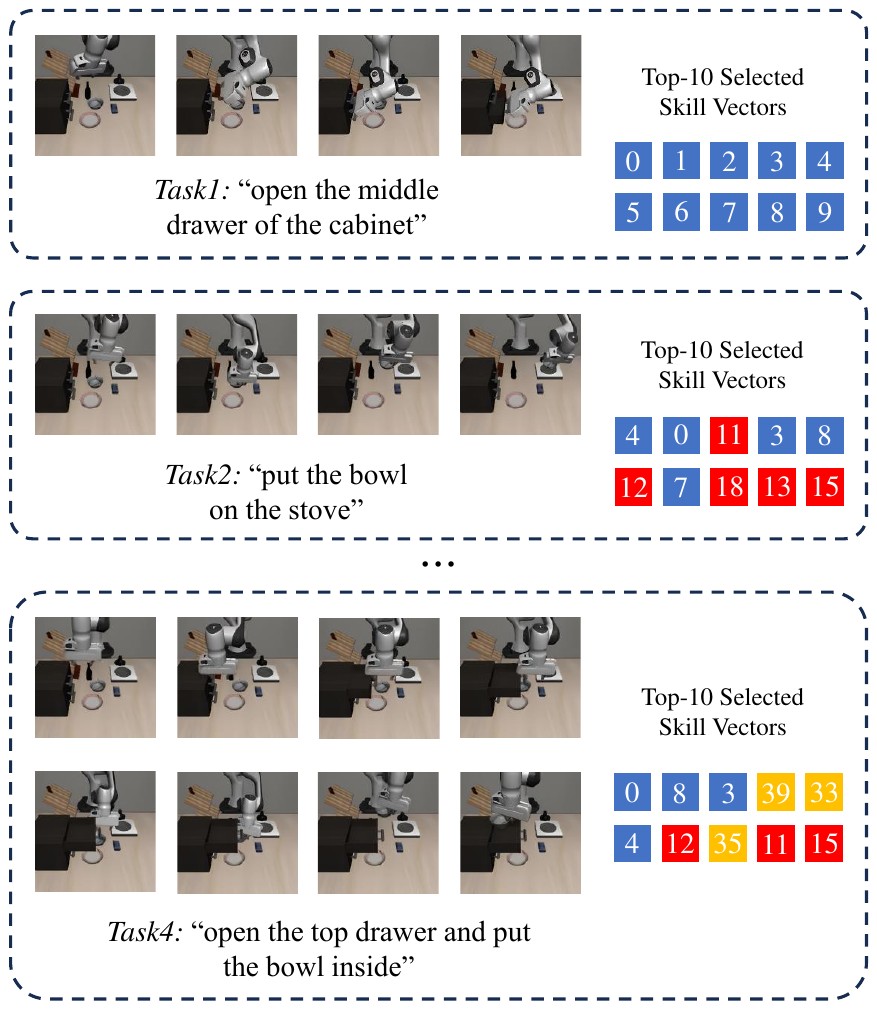}
\caption{Visualization of skill vector cross-task reuse in our SPECI under PackNet paradigm after learning task 4 on LIBERO-GOAL. The top-10 most frequently selected skills are displayed with color indicating their source skill subset and numerical labels showing their original indices.}
\label{skill_prompts_selection}
\end{figure}

Fig. \ref{er_packnet_each_task_suc} illustrates that the proposed \ac{SPECI}, when integrated with the ER \cite{chaudhry2019tiny} strategy, consistently achieves the highest success rates across all stages of continual learning on LIBERO-OBJECT, with about 20\%-50\% performance improvements on tasks 1, 5, 7, and 8. Notably, zero-shot performance is observed on tasks 3, 4, 6, and 9 with success rates of 25\%, 30\%, 5\%, and 30\%, respectively. This highlights the critical role of the skill codebook and the preservation of task-sharing knowledge in facilitating forward knowledge transfer, which is crucial for generalization to unseen tasks. 
Furthermore, \ac{SPECI} demonstrates significantly greater stability compared to baseline methods utilizing the PACKNET \cite{mallya2018packnet} learning paradigm, particularly in tasks 6 and 9, thereby underscoring its robust resistance to catastrophic forgetting. 
Additionally, the proposed framework allows robotic agents to engage in backward knowledge transfer with PACKNET \cite{mallya2018packnet}, enabling them to leverage shared task characteristics to enhance performance on previously learned tasks after encountering new ones. 
For instance, after learning task 9, the success rate for task 2 significantly improves relative to other comparison methods, as both tasks involve similar sub-tasks, specifically, picking up the bowl and placing it in a different location. Furthermore, learning tasks 5 (\textit{put the bowl on top of the cabinet}) and 6 (\textit{push the plate to the front of the stove}) contributes positively to task 4 (\textit{open the top drawer and place the bowl inside}), as task 4 involves a combination of pushing, pulling, and pick\&place actions, which are precisely the skill components represented in tasks 5 and 6.

We adopt PACKNET \cite{mallya2018packnet} as the core parameter update mechanism, primarily due to its intrinsic compatibility with our proposed \ac{SPECI} methodology through shared principles of parameter allocation. As quantitatively demonstrated in Table \ref{tab_lifelong_algor}, \ac{SPECI} achieves state-of-the-art bidirectional knowledge transfer compared to existing benchmark architectures across four distinct task suites. Specifically, our method demonstrates an average improvement of 17\% in \ac{FWT} across three short-horizon task suites, with a significant enhancement of 25\% observed on LIBERO-LONG. This empirical evidence substantiates the framework's exceptional capacity for forward knowledge transfer. 
Remarkably, our approach consistently yields negative scores in \ac{NBT} across all task suites, indicating that successive learning phases systematically enhance performance on previously learned tasks through validated skill reuse mechanisms. Furthermore, our framework outperforms competing methods by 17\% in \ac{AUC} for short-horizon tasks, with a remarkable margin of 27\% superiority on LIBERO-LONG. These consistent and wide performance margins across multiple evaluation metrics highlight the comprehensive superiority of our methodology in complex, long-horizon task settings.

\begin{figure}[ht]
    \centering
\subfloat{\label{subfig:fwt}\includegraphics[width=0.49\textwidth]{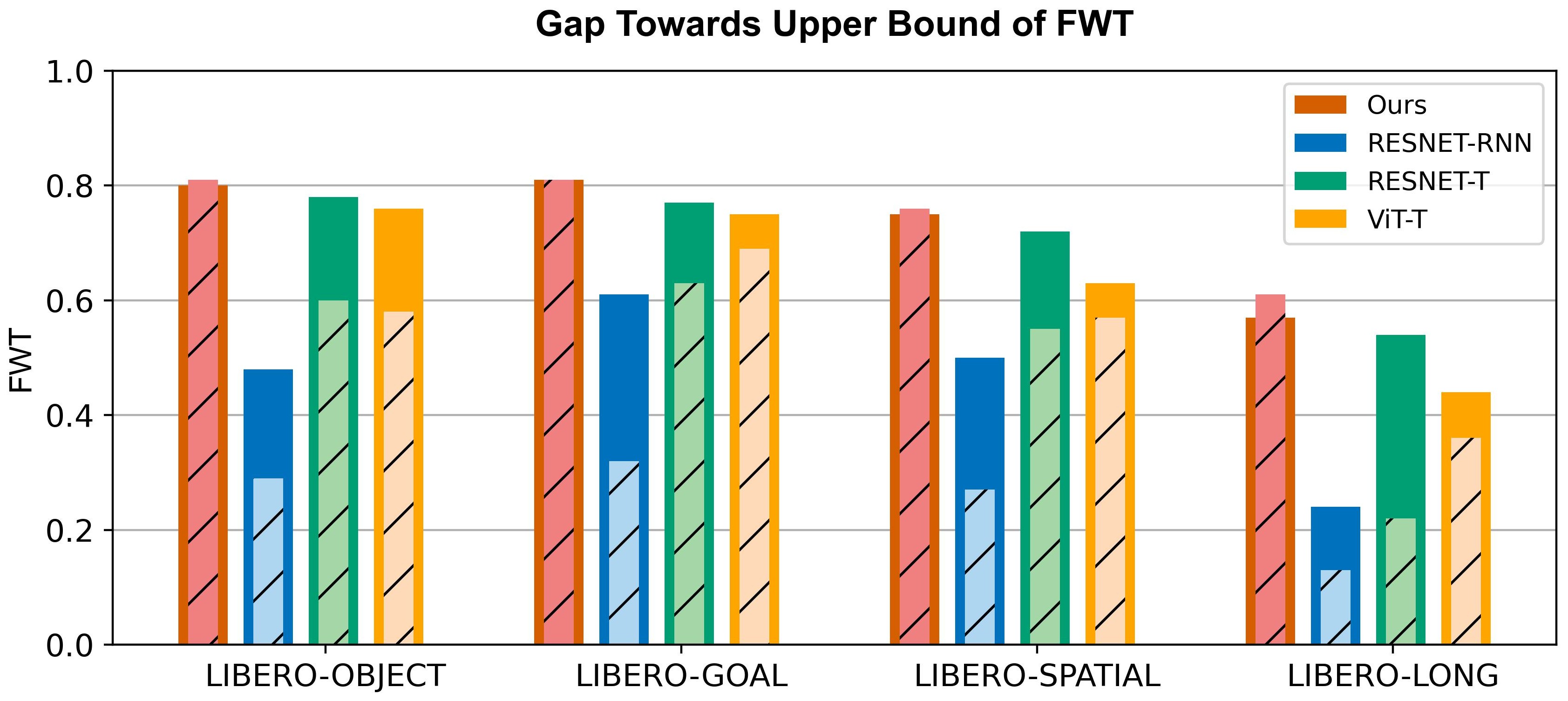}} \\
    \vspace{-2mm} 
    \subfloat{\label{subfig:auc}\includegraphics[width=0.49\textwidth]{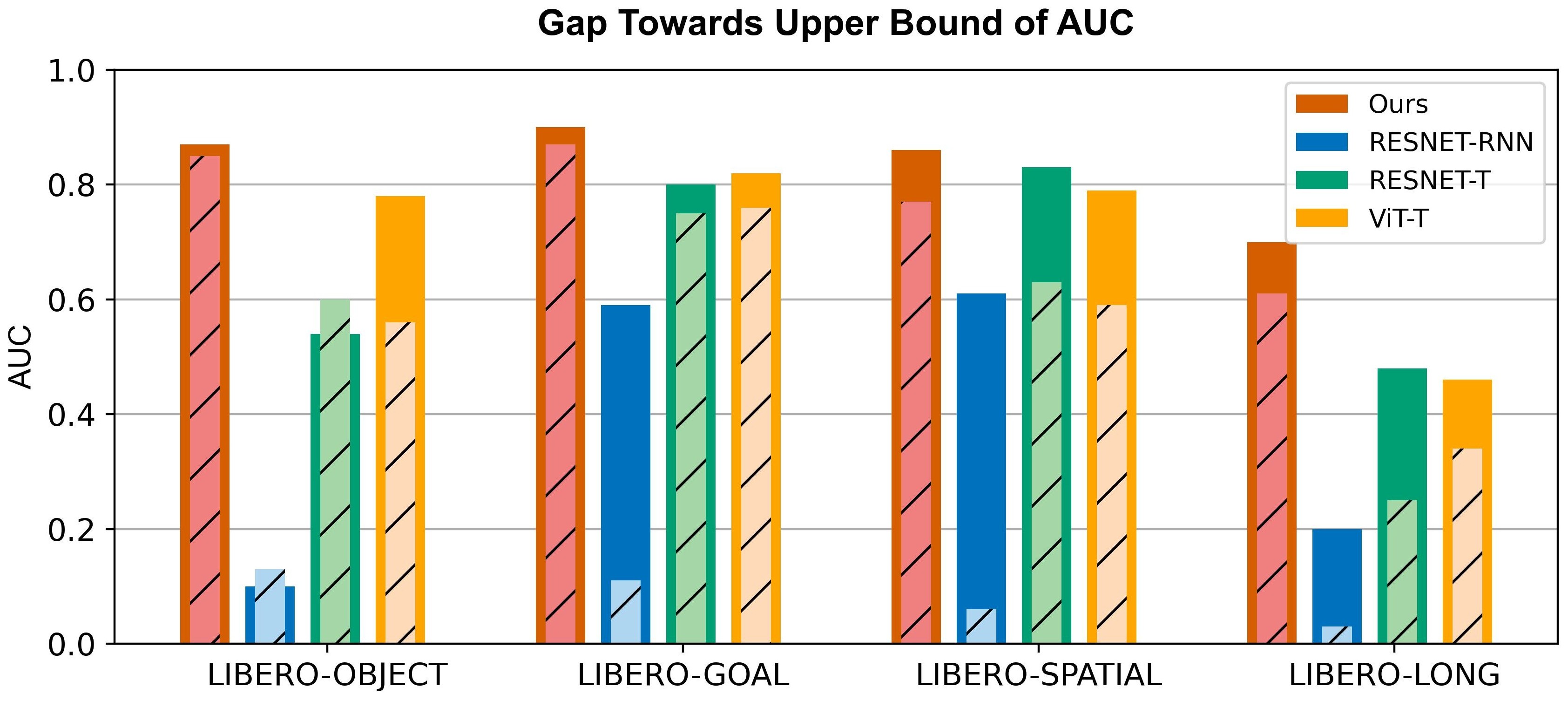}}
    \centering
    \caption{Visualization of the FWT and AUC metric gaps between the upper bounds (wide solid bars) and different policy architectures under PackNet paradigm (narrow hatched bars), evaluated across four task suites.}
    \label{mtl_seql_upper_bound}
\end{figure}

Fig. \ref{skill_prompts_selection} illustrates the task execution workflow of \ac{SPECI} under PACKNET \cite{mallya2018packnet} learning paradigm on LIBERO-GOAL task suite, along with the top-10 frequently selected skill vectors recorded. Notably, the blue-coded skills for task1 exhibit cross-task reutilization in task2, attributed to their shared rightward motion requirements. In contrast, the composite objective of task4 (drawer opening and bowl placement) triggers frequent activation of skill vectors from both task1 and task2 (red-coded). These findings validate \ac{SPECI}'s inherent ability for skill-level forward knowledge transfer, achieved through its dynamic skill codebook expansion and attention-guided skill sharing mechanism, which effectively enhance rapid adaptation to novel tasks.

\begin{table*}[htbp]
    \setlength{\tabcolsep}{7.5pt}
    \centering
    \caption{Ablation evaluation results of \ac{SPECI} under PackNet learning paradigm on LIBERO-OBJECT and LIBERO-GOAL task suites.}
    \label{tab_ablation}
    \begin{tabular}{lccccccc}
        \toprule
        \textbf{Methods} & & \multicolumn{3}{c}{\textbf{LIBERO-OBJECT}} & \multicolumn{3}{c}{\textbf{LIBERO-GOAL}} \\
        \cmidrule(lr){2-5} \cmidrule(lr){6-8}
        & & \textbf{FWT ($\uparrow$)} & \textbf{NBT ($\downarrow$)} & \textbf{AUC ($\uparrow$)} & \textbf{FWT ($\uparrow$)} & \textbf{NBT ($\downarrow$)} & \textbf{AUC ($\uparrow$)} \\
        \midrule
        \textbf{ResNet-T} & & $0.60 \pm 0.07$ & $0.17 \pm 0.05$ & $0.60 \pm 0.05$ & $0.63 \pm 0.02$ & $0.06 \pm 0.01$ & $0.75 \pm 0.01$ \\
        \textbf{ResNet-T w. M.A.} & & $0.76 \pm 0.01$ & $0.07 \pm 0.01$ & $0.74 \pm 0.01$ & $0.71 \pm 0.03$ & $0.05 \pm 0.00$ & $0.73 \pm 0.01$ \\
        \textbf{ResNet-T w. Codebook} & & $0.71 \pm 0.00$ & $0.04 \pm 0.01$ & $0.72 \pm 0.00$ & $0.75 \pm 0.01$ & $0.01 \pm 0.00$ & $0.82 \pm 0.02$ \\
        \textbf{ResNet-T w. Hier.} & & $0.77 \pm 0.01$ & $0.01 \pm 0.01$ & $0.79 \pm 0.02$ & $0.77 \pm 0.01$ & $0.01 \pm 0.01$ & $0.82 \pm 0.00$ \\
        \textbf{SPECI (Ours)} & & \textbf{0.81} $\pm$ \textbf{0.03} & \textbf{-0.01} $\pm$ \textbf{0.02} & \textbf{0.85} $\pm$ \textbf{0.03} & \textbf{0.81} $\pm$ \textbf{0.01} & \textbf{-0.01} $\pm$ \textbf{0.00} & \textbf{0.87} $\pm$ \textbf{0.01} \\
        \bottomrule
    \end{tabular}
\end{table*}

As illustrated in Fig. \ref{mtl_seql_upper_bound}, \ac{SPECI} achieves superior upper bounds in both FWT and AUC metrics across all task suites, while maintaining minimal deviation from the upper bounds in the continual learning setting. In terms of FWT, our method achieves competitive results against the upper bounds on LIBERO-OBJECT, LIBERO-GOAL, and LIBERO-SPATIAL task suites. This contrasts sharply with the substantial average performance gaps of 18.3\%, 16.3\%, and 15.3\% observed in other strong baselines. These quantitative comparisons substantiate our framework's strong forward knowledge transfer capability and learning efficiency compared to full-finetuning approaches. Notably, \ac{SPECI} yields an exceptional 4\% performance surplus beyond the upper bound on LIBERO-LONG, indicating that complex task sequences may introduce significant interference in sequential learning. 
This counterintuitive phenomenon can be attributed to our task-specific parameterization strategy, which effectively captures distinctive task representations while enhancing knowledge consolidation efficiency during long-horizon task execution. 
In terms of AUC, our framework exhibits only a 5.8\% average performance degradation throughout lifelong learning iterations, compared to a sharp 22.7\% decline observed in baseline methods. It's worth noting that \ac{SPECI} maintains robust performance on the challenging LIBERO-LONG benchmark, showing just 9\% degradation while competing methods suffer significant performance drops. These results validate the efficacy of our approach, which effectively integrates implicit skill modeling with systematic knowledge isolation and sharing mechanisms for long-horizon task learning.

\subsection{Ablation Experiments}
\label{Ablation Experiments}

To thoroughly evaluate the contribution of each component in our framework, we conduct ablation experiments on LIBERO-OBJECT and LIBERO-GOAL by incrementally integrating each module of \ac{SPECI} into the RESNET-T baseline. Specifically, RESNET-T w. M.A. incorporates the \textit{Mode Approximation} mechanism, RESNET-T w. Codebook integrates the dynamic skill codebook and the skill selection mechanism, RESNET-T w. Hier. simply duplicates the transformer decoder. As shown in Table \ref{tab_ablation}, the \textit{Mode Approximation} mechanism significantly improves \ac{FWT} through task-specific parameterization, while greatly reducing \ac{NBT} from 0.17 to 0.07 on LIBERO-OBJECT. This confirms its effectiveness in preserving task-sharing parameters for similar tasks. Although task-specific parameters are employed to accommodate emerging tasks, the slightly suboptimal \ac{AUC} performance on LIBERO-GOAL shows its limitation for tasks with different goals, highlighting the necessity for fine-grained skill knowledge preservation and sharing. The skill codebook design delivers balanced improvements, boosting \ac{FWT} by 11.5\%, reducing \ac{NBT} by 9\%, and increasing \ac{AUC} by 9.5\%. These results demonstrate that structured skill acquisition and preservation substantially enhance bidirectional knowledge transfer. Meanwhile, the hierarchical architecture improves overall performance through planning-execution decoupling and expanded model capacity. Ultimately, the complete \ac{SPECI} framework achieves optimal performance while maintaining negative \ac{NBT}, demonstrating its ability to leverage future knowledge for refining past task performance.

\section{Conclusion}
In this paper, we propose \ac{SPECI}, a novel end-to-end hierarchical continual learning framework  for robot manipulation. \ac{SPECI} achieves superior performance in complex long-horizon tasks  through three key components: a \textit{Multimodal Perception and Fusion Module} for environment comprehension and cross-modal information integration, a \textit{Skill Inference Module} featuring  dynamic codebook expansion for autonomous skill acquisition, and an \textit{Action Execution Module} utilizing \ac{GMM}-based policy to generate diverse action sequences. 
\ac{SPECI} establishes a unified framework for procedural and declarative knowledge transfer through 1) its self-evolving skill codebook that eliminates manual skill abstraction, 2) its attention-based key-query mechanisms that enables adaptive skill composition, and 3) its \textit{Mode Approximation} mechanism that facilitates task-level knowledge transfer. Extensive experimental evaluations demonstrate that \ac{SPECI} achieves significant improvements over state-of-the-art methods in both forward and backward knowledge transfer and overall performance.

While \ac{SPECI} demonstrates strong performance, its current implementation relies on ER or PACKNET paradigms which require explicit data replay or task identifiers. Future research will investigate rehearsal-free CL paradigms by integrating vision-language models (VLMs), enabling autonomous task-goal inference to enhance generalization in open-world scenarios. Furthermore, \ac{SPECI}'s motion reasoning architecture could be extended to a more granular hierarchical framework. A promising direction involves introducing task decomposition layers capable of translating high-level instructions into structured sub-task sequences before skill-based execution, thereby bridging the gap between abstract commands and low-level robotic actions.

\bibliographystyle{IEEEtran}
\bibliography{IEEEabrv, ref}


\end{document}

%% file: acro.tex
\newacro{SOTA}[SOTA]{state-of-the-art}
\newacro{RNN}[RNN]{Recurrent Neural Network}
\newacro{MAE}[MAE]{\textit{Mean Absolute Error}}
\newacro{ACC}[ACC]{\textit{Accuracy}}
\newacro{CNN}[CNN]{Convolutional Neural Network}
\newacro{MSE}[MSE]{Mean Square Error}
\newacro{LSTM}[LSTM]{Long Short-Term Memory}
\newacro{MLP}[MLP]{Multi-Layer Perception}
\newacro{DGR}[DGR]{Deep Generative Replay}
\newacro{IL}[IL]{Imitation Learning}
\newacro{RL}[RL]{Reinforcement Learning}
\newacro{BC}[BC]{Behavioral Cloning}
\newacro{IRL}[IRL]{Inverse Reinforcement Learning}
\newacro{SPECI}[SPECI]{\underline{\textbf{S}}kill \underline{\textbf{P}}rompts based Hi\underline{\textbf{E}}rarchical \underline{\textbf{C}}ontinual \underline{\textbf{I}}mitation Learning}
\newacro{MSA}[MSA]{multi-head self-attention}
\newacro{MCA}[MCA]{multi-head cross-attention}
\newacro{CIL}[CIL]{Continual Imitation Learning}
\newacro{GMM}[GMM]{Gaussian Mixture Model}
\newacro{CP}[CP]{CANDECOMP/PARAFAC}
\newacro{FWT}[FWT]{Forward Transfer}
\newacro{NBT}[NBT]{Negative Backward Transfer}
\newacro{AUC}[AUC]{Area Under the Success Rate Curve}
\newacro{CL}[CL]{Continual Learning}